\pdfoutput=1
\documentclass[11pt]{article}

\usepackage[final]{mtsummit25}
\usepackage{copyright}
\usepackage[T1]{fontenc}
\usepackage{times}
\usepackage{latexsym}
\usepackage[utf8]{inputenc}
\usepackage{hyperref}       % hyperlinks
\usepackage{url}            % simple URL typesetting
\usepackage{booktabs}       % professional-quality tables
\usepackage{amsfonts}       % blackboard math symbols
\usepackage{nicefrac}       % compact symbols for 1/2, etc.
\usepackage{microtype}      % microtypography
\usepackage{lipsum}
\usepackage{fancyhdr}       % header
\usepackage{graphicx}       % graphics
\graphicspath{{media/}}     % organize your images and other figures under media/ folder
\usepackage{arydshln}
\usepackage{verbatim}
\usepackage{listings}
\usepackage{multirow}
\title{
Instruction-tuned Large Language Models for Machine Translation\\ in the Medical Domain 
}

\author{
   Miguel Rios \\
  Centre for Translation Studies, University of Vienna \\
  \texttt{miguel.angel.rios.gaona@univie.ac.at} 
}

\begin{document}
\maketitle

\begin{abstract}
%\lipsum[1]
Large Language Models (LLMs) have shown promising results in machine translation, particularly for high-resource settings. However, in specialised domains such as medicine, their translation quality underperforms compared to standard Neural Machine Translation models, particularly regarding terminology consistency. In this study, we investigate the impact of instruction tuning for enhancing LLM performance in machine translation for the medical domain. We compare baseline LLMs with instruction-tuned models, and explore the impact of incorporating specialised medical terminology into instruction-formatted fine-tuning datasets. Our results show that instruction tuning significantly improves LLM performance according to automatic metrics. Furthermore, error analysis based on automatic annotation shows a substantial reduction in translation errors in the instruction-tuned models compared to the baselines.

\end{abstract}

\section{Introduction}

Current state-of-the-art Large Language Models have shown promising results in machine translation for high-resource language pairs and domains \citep{bawden-yvon-2023-investigating}. However, in low-resource domains (e.g. medical) LLMs have shown lower performance compared to standard neural machine translation (NMT) models \citep{bawden-yvon-2023-investigating,pourkamali2024machinetranslationlargelanguage}. The accuracy and consistency in the machine translation of terminology, syntax, and document structure is crucial for users, researchers, and translators who post-edit machine translated documents in high-risk domains \citep{Zakaryia_covid,pang2024saluteclassicrevisitingchallenges}. %gaona-etal-2023-quality
Moreover,  the introduction of in-domain translation constraints during generation into neural models is currently an open problem \citep{saunders-etal-2019-domain,alves2023steeringlargelanguagemodels,iikkaMittra_eatm24}. 

LLMs are trained to perform different Natural Language Processing (NLP) tasks such as summarisation, question answering, and translation, where users interact with the models via instructions (e.g. chat interface) \citep{touvron2023llama2openfoundation,openai2024gpt4technicalreport,dubey2024llama3herdmodels}. Instruction-tuning is a technique that leverages datasets from different NLP tasks, structured as prompts, for fine-tuning LLMs to enhance generalisation across novel tasks and domains \citep{chung2022FLANT5}. For example, in machine translation (MT), translating a segment from the European Medicines Agency corpus with specialised terminology the prompt can be framed as: "\textit{Glossary: medicine -> medicamento. \\ Translate the source text from English to Spanish following the provided translation glossaries. \\ English: The medicine was effective in patients with all three types of homocystinuria. \\ Spanish: }".  

Moreover, \citet{Alves2024TowerAO} instruction-tuned Llama-2 \citep{touvron2023llama2openfoundation} to perform translation related tasks, such as  segment and document level translation, post-editing, terminology aware translation, and error annotation.  The controlled generation of MT output with the correct terminology, segment length, or syntax can be framed as an instruction-tuning task for LLMs. Thus, improving the workflow of translation during post-editing with an instruction-following (i.e. chat) interface for an LLM tuned on a specific domain.
 
We seek to answer the following research question: Does instruction-tuning based on terminology rules improve translation quality on LLMs?  In this paper, we show results for adding specialised medical dictionaries into fine-tuning for LLMs. In particular, we follow the methodology from \citep{Alves2024TowerAO} by incorporating terminology information into the instruction-tuning datasets. Unlike \citep{Alves2024TowerAO}, our approach relies on openly available medical dictionaries and employs simple heuristics to construct instruction-tuning datasets.   An instruction-based interface could facilitate the interaction between professional translators and LLMs, and enables model customisation via the integration with user-defined terminology dictionaries.
%\newpage

Our contributions are as follows: We use parameter-efficient fine-tuning (PEFT) and quantisation of large language models (LLMs) for in-domain translation. We leverage  medical dictionary term pairs with parallel data to construct prompts that guide LLMs in translating specific terminology.

We evaluate FLAN-T5 \citep{chung2022FLANT5}, Llama-3 \citep{dubey2024llama3herdmodels}, and Tower \citep{Alves2024TowerAO} for English-Spanish, English-German, and English-Romanian language pairs in a split of a medical domain dataset. 

The instruction-tuned models outperformed the baseline models with the automatic metrics BLEU \citep{papineni-etal-2002-bleu}, chrF \citep{popovic-2015-chrf}, and COMET \citep{rei-etal-2020-comet}. Moreover, instruction-tuning improves the overall accuracy of the terminology. Finally, we evaluate the two best models with automatic error annotation \citep{guerreiro-etal-2024-xcomet}, and quality estimation \cite{rei-etal-2023-QEscaling}.     

\section{Background and Related Work}
\label{background}
 
Auto-regressive language models predict the next token in a sequence given a prefix context \citep{jelinek98, nnlm}, where LLMs are pre-trained with large amounts of texts followed by fine-tuning on different downstream tasks \citep{openai2024gpt4technicalreport}. In addition, \citet{chung2022FLANT5} propose to fine-tune LLMs with a mixture of several NLP datasets into an instruction format to improve: generalisation to unseen tasks, and generation given instruction prompts. For a machine translation task, the LLM is conditioned on a user defined prompt that consists of a translation instruction along with the source text to be translated \citep{pang2024saluteclassicrevisitingchallenges}. During testing, zero-shot prompting involves querying an LLM with a test input that was not present in the training data. For example, MT instruction includes a prompt asking to translate from a source language to a target language, and the corresponding source text. However, few-shot prompting provides a few examples of the translation task along with the test input to guide the LLM generation. In MT, few-shot examples consist of parallel source and human-translated sentences.

Supervised fine-tuning (SFT) is one of the most popular techniques for domain adaptation in LLMs, where models continue their training with a sample of in-domain data \citep{alves2023steeringlargelanguagemodels,sapEAMT2024}. However, SFT for LLMs  requires large amounts of computational resources, given that during training models update billions of parameters.  The goal of Parameter-efficient fine-tuning (PEFT) is to update (i.e. tune) a minimal set of parameters to achieve a similar performance compared to full SFT on downstream tasks. \citet{hu2021loralowrankadaptationlarge} propose low-Rank adaptation (LoRA) that freezes all the pre-trained model parameters and adds adapter trainable low-rank decomposition matrices of parameters into each layer of the model. 

Moreover, \citet{dettmers2023qloraefficientfinetuningquantized} propose that during fine-tuning to quantise the parameters of the pre-trained model into fewer bits (e.g. 4-bit) and keep the LoRA adapters with standard precision, thus reducing the memory usage. PEFT and quantisation with QLoRA enables academic translation practitioners to fine-tune LLMs with limited computing resources.  Llama versions 2 and 3 \citep{touvron2023llama2openfoundation, dubey2024llama3herdmodels} are open-source LLMs with different parameter scales, which are instruction-tuned for multiple Natural Language Processing tasks. Moreover, Llama has become the base model for the MT related work \citep{alves2023steeringlargelanguagemodels, pang2024saluteclassicrevisitingchallenges, sapEAMT2024}.

%\section{Related Work}

\citet{zhang-etal-2023-machine} compared 15 baseline LLMs and fine-tuned with QLoRA on different MT tasks (e.g. segment and document level translation) for the French-English language pair. Llama-2 outperformed other LLMs, fine-tuning improves performance on models that struggle on a few-shot setup, and QLoRA is potentially superior to full SFT in terms of efficiency. \citet{alves2023steeringlargelanguagemodels} compared instruction tuning with LoRA to few-shot prompting using Llama-2 in various language pairs. Fine-tuning outperforms the few-shot learning, is comparable to full SFT, requires few training data, and tackles over generation. However, LLMs struggle with translation directions out of English (en-xx).  \citet{Alves2024TowerAO} proposed Tower with a focus on translation related tasks, for example, document level translation, post-editing, and terminology-aware prompts. Tower is based on the continued training of Llama-2 with parallel translation data, and is followed by instruction-tuning for the MT tasks.   

\citet{zheng2024finetuninglargelanguagemodels} proposed to fine-tune LLMs based on prompts, and compared it to LoRA for domain adaptation in IT for Chinese-English and English-Chinese MT. Moreover, \citet{zheng2024finetuninglargelanguagemodels} incorporate IT terminology by few-shot prompting and chain-of-thought. The template used for the proposed prompt-tuning model has a substantial impact on performance, and the introduction of terminology with simple prompt rephrasing outperforms chain-of-thought.  \citet{sapEAMT2024} studied domain adaptation for business IT with LLMs. They compared full SFT, LoRA, different prompting techniques, and standard NMT. Finally, \citet{sapEAMT2024} defined guidelines for domain adaptation with LLMs.       
\citet{moslem-etal-2023-adaptive} evaluate LLMs for translation on specialised domains (e.g. medical COVID-19), and incorporate terms from glossaries into their prompts to tackle issues with no retrieved matches in few-shot learning. \citet{jerpelea-etal-2025-dialectal} developed a parallel dataset for the low-resource languages Romanian, and Aromanian, they instruction-tuned Llama-3 for Romanian, and compared multilingual NMT, GPT, Llama-3, and Tower for translation. 

We followed \citep{alves2023steeringlargelanguagemodels,Alves2024TowerAO} for our experimental design. Unlike previous work on LLM for MT, our approach focuses on the medical domain, relies on openly available medical dictionaries, employs simple heuristics to construct the instruction-tuning datasets, and uses efficient tuning techniques. In particular, we evaluate the impact of instruction tuning for improving terminology translation in LLMs. 
%We establish baseline performance by evaluating FLAN-T5, Llama-3, and Tower. Subsequently, we instruction-tune these models using QLoRA and incorporating terminology-aware prompts.

\section{Experimental Setup}
\label{results}
\subsection{Data}

We use the corpus of the European Medicines Agency (EMEA) \citep{elrc30_emea} for the English-Spanish (en-es), English-German (en-de), and English-Romanian (en-ro) language pairs. The EMEA corpus contains multilingual PDF documents from the European Medicines Agency,  automatically converted to text and aligned at the segment level. We randomly split the EMEA corpus into 20K segments for each language pair. These subsets were then merged into a single tuning dataset of 60K segments. Furthermore, we created separate validation and test sets, each containing 500 segments per language pair.

\subsection{Terminology Annotation}
The Interactive Terminology for Europe (IATE)\footnote{\url{https://iate.europa.eu/download-iate}} is a terminology management system from EU institutions that covers different domains (e.g. economics, law, health). For our source and target language pairs, we downloaded the IATE database in the \textit{health} domain (\textit{id 2841}).  We only used terms with quality 3 (reliable) and 4 (very reliable) stars (human annotated quality scores), resulting in 38,898 terms for en-es, 49,828 terms for en-de, and 9,551 terms for en-ro. 

We incorporate medical terms as translation instructions by identifying term pairs within each aligned segment. For every aligned segment, we retrieve candidate terms using strict matching, which requires the presence of a candidate pair in both the source and target segments. If one or more candidate pairs are identified, we include them in the instruction template within the prompt. For example, an instruction-tuning input in en-es "\textit{spectrum of activity -> espectro de actividad, amoxicillin -> amoxicilina, and activity -> actividad}" are term pairs identified in the parallel segment:

%\newpage
\begin{lstlisting}
 Glossaries:
 "spectrum of activity" -> "espectro de actividad"
 "amoxicillin" -> "amoxicilina"
 "activity" -> "actividad"
 Translate the source text from English to Spanish following the provided translation glossaries.
 English: Amoxicillin is susceptible to degradation by beta-lactamases produced by resistant bacteria and therefore the spectrum of activity of amoxicillin alone does not include organisms which produce these enzymes.
 Spanish: La amoxicilina es sensible a la degradación por las beta-lactamasas producidas por bacterias resistentes y por tanto el espectro de actividad de la amoxicilina sola no incluye microorganismos productores de estas enzimas.   
\end{lstlisting}

When no candidates are identified within a segment, the instruction consists only of the translation task prompt. For example, an instruction-tuning input in en-es:
%\newpage
\begin{lstlisting}
Translate the source text from English to Spanish.
English: Do not use Cymevene if you are breast-feeding.
Spanish: No use Cymevene si está en periodo de lactancia.
\end{lstlisting}

The prompt templates for the baseline models are defined in Section \ref{inst_templates}, and  we perform zero-shot prompting to generate translations for en-es, en-de, and en-ro. For example, a test input in en-es:
%\newpage
\begin{lstlisting}
Glossary:
"insulin" -> "insulina"
Translate the source text from English to Spanish following the provided translation glossaries.
English: Within-subject variability of the time action profile of Levemir and NPH insulin Pharmacodynamic Endpoint
Spanish:
\end{lstlisting}

\subsection{Models}

We use the HuggingFace transformers framework for the baseline LLMs \citep{wolf2020huggingfacestransformersstateoftheartnatural}, and PEFT \citep{peft} for the instruction-tuning with QLoRA. Our baseline LLMs are as follows: FLAN-T5-large\footnote{\url{google/flan-t5-large}} (783M parameters), an encoder-decoder model; Llama-3-8B\footnote{\url{meta-llama/Meta-Llama-3.1-8B-Instruct}}, an instruction-tuned LLM for NLP tasks; and Tower-7B\footnote{\url{Unbabel/TowerInstruct-7B-v0.2}}, an instruction-tuned LLM for MT tasks. We evaluated two distinct instruction-tuned model architectures: encoder-decoder model based on FLAN-T5, and auto-regressive LLMs based on Llama. 

We use QLoRA with a 4-bit quantisation to fine-tune each baseline model for one epoch on the tuning dataset (60K segments). The values for QLoRA and tuning hyper-parameters for each model are defined in Section \ref{hyper}, for FLAN-T5 \ref{tab:flan_param}, Llama-3-8B \ref{tab:llama_param}, and Tower-7B \ref{tab:tower_param}. Finally, for generation, we use zero-shot prompting and stochastic decoding with top-$p$ sampling $p=0.9$.  We release our scripts and data on GitHub at: \url{https://github.com/HAITrans-lab/instruction-tuned-medical-LLM} %\url{https://github.com/HAITrans-lab/instruction-tuned-medical-LLM}.

\section{Results}

We show results with automatic metrics and terminology accuracy for FLAN-T5, Llama-3 and Tower for the en-es, en-de and en-ro language pairs. Moreover, we show automatic error annotation, and quality estimation scores for the best performing models.  

\subsection{Automatic Metrics}

We evaluated all models with BLEU, chrF, and COMET in the test split. Table \ref{tab:scores} shows the comparison between the baselines and the instruction-tuned models with QLoRA. The BLEU, chrF, and COMET scores for the instruction-tuned models are statistically significant ($p < 0.05$) for all models.

\begin{table*}[ht!]
    \centering
     \resizebox{\textwidth}{!}{%
    \begin{tabular}{lccccccccc}
        \toprule
        Model &  & en-es$\uparrow$ &  &  & en-de$\uparrow$ &  &  & en-ro$\uparrow$  & \\ \cmidrule(lr){2-4}\cmidrule(lr){5-7}\cmidrule(lr){8-10}
        
         & BLEU & chrF & COMET & BLEU & chrF & COMET & BLEU & chrF & COMET \\
         \midrule
         FLAN-T5 & 28.51 & 57.11 & 0.73 & 14.76 & 43.86 & 0.63 & 17.34 & 45.00 & 0.64 \\
         QLoRA FLAN-T5 & 36.43 & 63.40 & 0.78 & 25.45 & 54.93 & 0.72 & 28.65 & 57.44 &  0.77 \\ \hdashline 
         %Llama3-8B & 14.90 & 52.14 & 0.52  & 11.70 & 49.91 & 0.52 & 14.07 &  48.30 & 0.55 \\
         %fine-tuned Llama3-8B & 20.47 & 56.87 &  0.66 & 16.95 & 53.96 & 0.64 & 22.56 & 55.50  & 0.69 \\ \hdashline
         Llama-3-8B & 34.07 & 63.02 & 0.79  & 25.44 & 58.08  & 0.78 & 24.99  &  53.17 & 0.76 \\
         QLoRA Llama-3-8B & 45.07 & 67.74 & 0.85 & 36.30 & 62.21 & 0.84 & \textbf{35.97} & \textbf{61.19}  & \textbf{0.85} \\
         \hdashline 
         Tower-7B & 42.27 & 66.31 &  0.86 & 34.80 & 62.45 & 0.85 & 18.20  & 44.86  & 0.69 \\
         QLoRA Tower-7B & \textbf{48.88} & \textbf{70.36} & \textbf{0.87}  & \textbf{42.11} & \textbf{67.62} & \textbf{0.87} & 23.93 & 50.57  & 0.78 \\
         \bottomrule
    \end{tabular}
    }
    \caption{Comparing the baseline and QLoRA fine-tuned LLMs with \textbf{automatic metrics} for the en-es, en-de, and en-ro language pairs. }
    \label{tab:scores}
\end{table*}

To prevent over-generation and improve the performance of Llama-3, we post-processed the output by cutting it at the first appearance of the end-of-sequence token "\textit{<|eot\_id|>}". As noted by \citet{zhang-etal-2023-machine}, Llama models repeat the translation output or produce \textit{assistant} suggestions to improve the prompts along with the translation. 
 
In Table \ref{tab:scores},  Tower and the QLoRA Tower outperform the other models with the automatic metrics for en-es, and en-de. However, Romanian (en-ro) is not present in the original Tower fine-tuning for MT. Tower is based on LLaMA-2 which is not focused on multilingual data, in contrast to Llama-3. Moreover, QLoRA tuning produced improvements for all models.  

As shown in Table \ref{tab:scores}, Tower and QLoRA Tower achieved the highest automatic metric scores for en-es, and en-de. However, the original Tower model was not fine-tuned for en-ro MT. Furthermore, Tower is based on LLaMA-2, which is less focused on multilingual data compared to Llama-3. Nonetheless, the QLoRA models consistently improved performance across all models. The bold numbers are the best automatic scores across all models for a given language pair. 

\paragraph{Terminology Accuracy} We compute the accuracy of the terminology in the MT output compared to the reference translations. To compute accuracy, the exact term must be present in both the MT segment and the database to be correct. Table \ref{tab:accuracy} shows the accuracy scores for the terminology. Instruction-tuning improves the accuracy of terms across models, where Flan-T5 followed by Tower achieve the highest terminology accuracy performance. We observed that the LLM produced translations with increased terminology accuracy for the high-resource language pairs, en-de and en-es.

\begin{table}[h!]
    \centering
    \resizebox{0.95\columnwidth}{!}{%
    \begin{tabular}{lccc}
        \toprule
        Model &  en-es$\uparrow$ & en-de$\uparrow$ & en-ro$\uparrow$  \\ 
        %\cmidrule(lr){2-4}\cmidrule(lr){5-7}\cmidrule(lr){8-10}
        
         \midrule
         FLAN-T5 & 0.72 & 0.45 & 0.38 \\
         QLoRA FLAN-T5 & 0.90 &  \textbf{0.91} & \textbf{0.90} \\
         \hdashline
         Llama-3-8B & 0.59 & 0.53 & 0.44 \\
         QLoRA Llama-3-8B & 0.69 & 0.68 & 0.51 \\
         \hdashline 
         Tower-7B & 0.88 & 0.79 &  0.58 \\
         QLoRA Tower-7B & \textbf{0.91} & 0.86 &  0.68 \\
         \bottomrule
    \end{tabular}
    }
    \caption{Comparing the baseline and QLoRA fine-tuned LLMs with \textbf{terminology accuracy} for the en-es, en-de, and en-ro language pairs. }
    \label{tab:accuracy}
\end{table}

\subsection{Automatic Error Annotation}

Automatic metrics are not designed to identify specific translation errors in MT outputs, for example, errors in terminology. Multidimensional Quality Metrics (MQM) \cite{lommel_multidimensional_2014} are based on manually classifying and annotating errors using predefined categories. The MQM error typology covers high-level error categories (e.g. Accuracy, linguistic conventions, style, etc.), where each category can be further expanded into fine-grained categories (e.g. Accuracy into Mistranslation, addition, untranslated, etc.). Expert translators identify an error in the MT output, label it with a category from the typology, and also assign a severity score to it. The severity weights defined in \cite{freitag_results_2021} are: minor $\times$ 1 (MIN), major $\times$ 5 (MAJOR), and critical $\times$ 10 (CRIT). The MQM score  is defined as follows: 
\begin{equation}
\label{mqm_eq}
\small
  \resizebox{0.47\textwidth}{!}{$  \textrm{MQM} = 100 \cdot \left( 1- \frac{10 \cdot \textrm{critical} + 5 \cdot \textrm{major} + \textrm{minor}}{\textrm{tokens}}\right), $}
\end{equation}

We use XCOMET \citep{guerreiro-etal-2024-xcomet} to produce automatic MQM annotations. XCOMET only annotates the error spans in the MT output with severities \footnote{\url{Unbabel/XCOMET-XL}}, and the corresponding prediction confidence for each span. The automatic error annotation with XCOMET is based on an LLM that required a larger GPU than our available resources for execution. We run the XCOMET evaluations on CPU where the process is slow, thus we only evaluate the best two models based on the automatic metrics, Llama-3 and Tower.  

We show the number of errors $\downarrow$ in Table \ref{tab:mqm} and the MQM scores $\uparrow$ in Table \ref{tab:mqmscores} for each system. The MQM score summarises the individual errors into a weighted score based on severity (Equation \ref{mqm_eq}). 
Table \ref{tab:mqm} shows the number or errors by severity for each model. The total number of errors for the instruction-tuned Llama-3 is: 1914 (en-es), 2910 (en-de), and 1764 (en-ro). The instruction-tuned Tower is: 745 (en-es), 1059 (en-de), and 1632 (en-ro). Instruction-tuned Tower shows fewer critical errors compared to Llama for the three language pairs (en-es, en-de, and en-ro).

\begin{table*}[t!]
    \centering
    \resizebox{0.89\textwidth}{!}{%
    \begin{tabular}{lccccccccc}
        \toprule
        Model &  & en-es$\downarrow$ &  &  & en-de$\downarrow$ &  &  & en-ro$\downarrow$  & \\ \cmidrule(lr){2-4}\cmidrule(lr){5-7}\cmidrule(lr){8-10}
        
         & MIN & MAJ & CRIT & MIN & MAJ & CRIT & MIN & MAJ & CRIT \\
         \midrule
         Llama-3-8B & 145 & 1277 & 1240  & 1693 & 719  & 938 & 95  & 983  & 1301 \\
         QLoRA Llama-3-8B & 359 & 1105 & 450 &  2160 &  295 & 455 & 225  & 844  & 695 \\
         \hdashline 
         Tower-7B &  592 & 241 & 15  & 1266 & 50  & \textbf{25} &  253 & 844 & 695 \\
         QLoRA Tower-7B & 583 & 149 & \textbf{13}  & 1007 &  26 & 26 & 503  & 868  & \textbf{261} \\
         \bottomrule
    \end{tabular}
    }
    \caption{Comparing the baseline and QLoRA fine-tuned LLMs with the number of \textbf{errors} with the following categories: minor (MIN), major (MAJ), and critical (CRIT). }
    \label{tab:mqm}
\end{table*}

\begin{table}[h!]
    \centering
    \resizebox{0.95\columnwidth}{!}{%
    \begin{tabular}{lccc}
        \toprule
        Model &  en-es$\uparrow$ & en-de$\uparrow$ & en-ro$\uparrow$  \\ 
        %\cmidrule(lr){2-4}\cmidrule(lr){5-7}\cmidrule(lr){8-10}
        
         \midrule      
         Llama-3-8B & 35.98 & 41.29 &  27.76 \\
         QLoRA Llama-3-8B & 58.83 & 59.45 & \textbf{45.66} \\
         \hdashline 
         Tower-7B & 82.35 & 80.70 & 20.11 \\
         QLoRA Tower-7B & \textbf{86.63} & \textbf{84.69} & 36.96 \\
         \bottomrule
    \end{tabular}
    }
    \caption{Comparing the baseline and QLoRA fine-tuned LLMs with \textbf{MQM scores} for the en-es, en-de, and en-ro language pairs. }
    \label{tab:mqmscores}
\end{table}

Table \ref{tab:mqmscores} presents the MQM scores, which show a reduction in critical, major, and minor errors after the instruction tuning phase. In these results, Tower outperforms Llama.

\paragraph{Automatic Error Annotation Analysis} We conducted a preliminary error analysis of the automatic error annotation for en-es to assess the quality of XCOMET to label translation errors. A native Spanish speaker with English proficiency served as the annotator. The limited number of examples analysed from the en-es automatic error annotation is because of the lack of a professional medical translator during the preliminary analysis. We show annotation examples between the baseline and instruction-tuned models for Llama-3 and Tower. 

Table \ref{tab:xcomet_eg} presents examples of automatic error annotations generated by XCOMET for Llama, QLoRA Llama, and Tower. For Llama, XCOMET identified a critical error with a confidence score of 0.52. Similarly, a critical error in the instruction-tuned Llama was annotated with a confidence of 0.40. While XCOMET produced incomplete annotations, potentially because of over-generation by Llama, it successfully identified code-switched words, such as "\textit{assistant}".

In Tower, XCOMET annotated a major error, "\textit{reconstitución,}" with a confidence of 0.50. For the instruction-tuned Tower, a minor error, "\textit{reconstitu,}" was annotated with a confidence of 0.42. Notably, "\textit{reconstitución}" is the correct term in the MT output with low prediction confidence. A potential solution involves filtering annotations based on a predefined confidence threshold, keeping only high-confidence predictions.

\begin{table*}[th!]
\centering
\resizebox{0.9\textwidth}{!}{
{\footnotesize
\begin{tabular}{p{2.5cm} p{8.5cm} p{3cm}}
\toprule
Terms                                                                                                                           & Source, reference, and MT                                                                                                                                                                                                                                                                                                                                                                                       & Annotation                                                                                                                                  \\ \hline
\multirow{3}{11em}{active substance -\textgreater principio activo, \\ system -\textgreater sistema, \\ fentanyl -\textgreater fentanilo}  & \textbf{src}: Ionsys transdermal system delivers the active substance, fentanyl, through the skin.                                                                                                                                                                                                                                                                                                  & \multirow{3}{10em}{\textbf{critical error}: assistant Here is the translation: Ionsys sistema transdérmico proporciona} \\ \cline{2-2}
                                                                                                                                & \textbf{ref}: El sistema de liberación transdérmica Ionsys administra el principio activo, el fentanilo, a través de la piel.                                                                                                                                                                                                                                                                       &                                                                                                                                             \\ \cline{2-2}
                                                                                                                                & \textbf{Llama}: Ionsys sistema transdérmico proporciona la sustancia activa, fentanilo, a través de la piel.\textbf{assistant  Here is the translation:   Ionsys sistema transdérmico proporciona} el principio activo, fentanilo, a través de la piel.  I corrected the translation using the provided glossary.assistant  Using the glossary "system" -\textgreater "sistema", I should have translated it to &                                                                                                                                             \\ \hline
\multirow{3}{10em}{active substance -\textgreater principio activo,  \\ system -\textgreater sistema, \\ fentanyl -\textgreater fentanilo} & \textbf{src}: Ionsys transdermal system delivers the active substance, fentanyl, through the skin.                                                                                                                                                                                                                                                                                                  & \multirow{3}{10em}{\textbf{critical error}: assistant El sistema transdérmico Ionsys libera el}                                  \\ \cline{2-2}
                                                                                                                                & \textbf{ref}: El sistema de liberación transdérmica Ionsys administra el principio activo, el fentanilo, a través de la piel.                                                                                                                                                                                                                                                                       &                                                                                                                                             \\ \cline{2-2}
                                                                                                                                & \textbf{QLoRA Llama}: El sistema transdérmico Ionsys libera el principio activo, el fentanilo, a través de la piel.  \textbf{assistant  El sistema transdérmico Ionsys libera el} principio activo, el fentanilo, a través de la piel.                                                                                                                                                                                &                                                                                                                                             \\ \hline
\multirow{3}{10em}{reconstitution -\textgreater reconstitución}                                                                    & \textbf{src}: Write the date of reconstitution and expiry on the label (expiry is 1 month after reconstitution)                                                                                                                                                                                                                                                                                     & \multirow{3}{10em}{\textbf{major error}: reconstitución}                                                                       \\ \cline{2-2}
                                                                                                                                & \textbf{ref}: Escriba la fecha de reconstitución y la de caducidad en la etiqueta (la caducidad es 1 mes después de la reconstitución)                                                                                                                                                                                                                                                              &                                                                                                                                             \\ \cline{2-2}
                                                                                                                                & \textbf{Tower}: Escriba la fecha de reconstitución y el de caducidad en la etiqueta (el de caducidad es 1 mes después de la \textbf{reconstitución}).                                                                                                                                                                                                                                                           &                                                                                                                                             \\ \bottomrule
\end{tabular}
}
}
\caption{Examples of \textbf{automatic error annotation} for en-es using XCOMET.}
\label{tab:xcomet_eg}
\end{table*}

\subsection{Quality Estimation}

Quality estimation (QE) models predict a quality score for the MT output without using reference translations. QE evaluation can be useful for cases of low-resource language pairs and practical applications, given the lack of reference translations. We use COMETKiwi \citep{rei-etal-2023-QEscaling} for QE evaluation \footnote{\url{Unbabel/wmt22-cometkiwi-da}}. COMETKiwi is based on COMET features to train a QE prediction model. The QE model is trained with an annotated multilingual source and corresponding MT outputs to predict quality based on direct assessment (i.e. ranking) or MQM scores.

\begin{table}[h!]
    \centering
    \resizebox{0.95\columnwidth}{!}{%
    \begin{tabular}{lccc}
        \toprule
        Model &  en-es$\uparrow$ & en-de$\uparrow$ & en-ro$\uparrow$  \\ 
        %\cmidrule(lr){2-4}\cmidrule(lr){5-7}\cmidrule(lr){8-10}
        
         \midrule      
         Llama-3-8B & 0.513 & 0.507 &  0.484 \\
         QLoRA Llama-3-8B & 0.657 & 0.619 &  0.595 \\
         \hdashline 
         Tower-7B & 0.840 & 0.806 &  0.647 \\
         QLoRA Tower-7B & \textbf{0.850} & \textbf{0.825} &  \textbf{0.754} \\
         \bottomrule
    \end{tabular}
    }
    \caption{Comparing the baseline and QLoRA fine-tuned LLMs with \textbf{QE} for the en-es, en-de, and en-ro language pairs. }
    \label{tab:qescores}
\end{table}
%Unbabel/wmt22-cometkiwi-da

Table \ref{tab:qescores} shows the comparison of QE scores for Llama-3 and Tower. The instruction-tuned Tower shows higher QE scores compared to Llama in all language pairs. The QE scores show a similar order in model quality compared to the output of automatic metrics without the need for reference translations.

\subsection{Discussion and Limitations}

Instruction-tuning improves the overall accuracy of terminology and translation quality (e.g. automatic metrics). Instruction-tuned FLAN-T5 (encoder-decoder) has the highest terminology accuracy, but its improvements in translation quality are lower compared to the LLMs.  A possible explanation is the difference in parameter size compared to the LLMs, and pre-trained data available for the LLMs. However, to achieve a more accurate evaluation, it is recommended to perform a manual error annotation with professional medical translators.

Both the baseline and instruction-tuned models generate terms defined by our prompts. However, fine-tuning substantially improves accuracy for FLAN-T5, Tower, and Llama-3. Furthermore, Tower includes terminology translation across diverse domains as a component of its tuning tasks.

Llama-3 presents over-generation, producing an excessive amount of tokens with assistant suggestions. For example, in en-es in the test set, the baseline model generates 29,569 tokens, which is reduced to 25,225 tokens after fine-tuning. Examples of this over-generation in Llama-3 include assistant-specific text alongside the expected machine translation output, such as: "\textit{..\{source segment\} assistant Here is the corrected translation: \{MT target segment\}...}". However, the instruction-tuned LlaMA-3 also over-generates: \textit{"..I corrected the translation using the provided glossary.assistant  Using the glossary..."},  or it continues repeating the MT output. A possible solution is to use a prompt that constrains the model to produce only the target segment. With our current prompt, both Llama-3 models require extra post-processing to extract the MT and avoid biases on the automatic metrics and automatic error annotation.  On the other hand, Tower generates 11,034 tokens compared to 10,906 tokens for the instruction-tuned. The MT tasks tuning on Tower improves translation accuracy and avoids over-generation.

However, given common limitations on academic computational resources (one GPU) we use small size LLMs (8B) with quantisation, PEFT for tuning our models, and a small split of the EMEA corpus. A limitation of quantisation is the use of pre-trained models with lower precision models that may hurt overall performance. However, SFT in LLMs can be achieved with significantly less data than training from scratch and other domain adaptation approaches \citep{zhu-etal-2024-fine}. The total size of the EMEA corpus is approximately 1M segments.

Automatic error annotation and QE scores offer a detailed evaluation of our language pairs and domain. However, XCOMET shows inaccuracies in terminology annotation, particularly with low-confidence predictions. Furthermore, to validate the reliability of automatic error annotation within the medical domain, a comprehensive analysis involving professional translators is essential. Additionally, XCOMET requires substantial GPU resources.

We use accuracy to evaluate the terms generated in the MT output. The limitation of accuracy is that context is not taken into account \citep{corral-saralegi-2024-morphology}, for example, translation quality is lowered with high term accuracy in FLAN-T5. A limitation of building our terminology prompt dataset using only exact matches is the potential to miss terms that are expressed differently depending on the context. Furthermore, the coverage of terms and domains within IATE represents a limitation of terminology databases. For example, in the parallel (en-es) segment:  \textit{"Posology for MDS/MPD The recommended dose of imatinib is 400 mg/day for adult patients with MDS/MPD."} and  \textit{"Posología para SMD/SMP La dosis recomendada de imatinib para pacientes adultos con SMD/SMP es de 400 mg/día"} from IATE the exact term match is "\textit{dose -> dosis}". However, IATE does not contain "\textit{MDS/MPD -> SMD/SMP}" that means "\textit{Myelodysplastic/Myeloproliferative Neoplasm}"\footnote{\url{https://www.cancer.gov/types/myeloproliferative/hp/mds-mpd-treatment-pdq}}. A possible solution is to combine translation terminology databases with medical ontologies, for example, Medical Subject Headings (MeSH)\footnote{\url{https://www.nlm.nih.gov/mesh/meshhome.html}}, and the Unified Medical Language System (UMLS)\footnote{\url{https://www.nlm.nih.gov/research/umls/index.html}} that has multilingual features.

\section{Conclusions and Future Work}

In this study, we show a comparison between baseline LLMs and QLoRA instruction-tuned models in the medical domain for en-es, en-de, and en-ro. We introduce medical terminology from IATE into an instruction-formatted dataset for controlled generation in LLMs. Instruction-tuned models significantly outperform the baseline across automatic evaluation metrics. Furthermore, these models show improved accuracy in terminology translation compared to the baseline.

In particular, the instruction-tuned Tower model presents superior translation quality according to different evaluation methods (automatic metrics, MQM annotation, and QE). Additionally, Tower requires fewer computational resources and less post-processing compared to LLaMA-3.

A limitation of our current evaluation is the reliance on automatic metrics and the limited quality of automatic error annotation. For future work, we will evaluate the baselines with few-shot instead of zero-shot. We will define different prompts for Llama-3 to avoid over-generation. We will perform an evaluation on a balanced test split in terms of the number and type of present terms with respect to the training data. Finally, we will perform a manual error annotation, as automatic metrics may not test for correct terminology generation on the MT output \citep{haque-etal-2019-investigating, gaona-etal-2023-quality}. %gaona-etal-2023-quality

\paragraph{Sustainability Statement} For the experiments we use a Tesla T4 GPU (16GB) from Azure with an approximate SFT time of 20 hours per model.  Instruction-tuning with PEFT tackles issues for scare computational resources (GPUs) for short training time (e.g. one epoch) and small tuning data (60K segments). Moreover, we performed automatic error annotation on the CPU instead of GPU given our academic computational limitations.

From \href{https://mlco2.github.io/impact#compute}{MachineLearning Impact calculator} presented in \citep{lacoste2019quantifying}: West-Europe Azure has a carbon efficiency of 0.57 kgCO$_2$eq/kWh. A cumulative of 100 hours of computation was performed on hardware of type T4 (TDP of 70W). Total emissions are estimated to be 3.99 kgCO$_2$eq of which 100 percent were directly offset by the cloud provider.

\section*{Acknowledgments}
This work was supported by the ZID of the University of Vienna with Azure cloud credits.

%Bibliography
%\bibliographystyle{plainnat}  
\bibliography{references} 

\appendix

\section{Hyper-parameters}
\label{hyper}

The hyper-parameter values tables for FLAN-T5, Llama-3-8B, and Tower-7B are as follows:

\begin{table}[h!]
    \centering
    \resizebox{0.57\columnwidth}{!}{
    \begin{tabular}{lc} \toprule
         Hyper-parameter & Value \\ \midrule
         r & 8\\
         $\alpha$ & 32\\
         Dropout & 0.1\\
         Target modules & q, v \\
         \midrule
         Max source length & 512\\
         Max target length & 512\\
         Batch size & 6\\
         Learning rate & $2e-4$\\
         Warm-up steps & 0.03\\
         Scheduler type & linear\\
         \bottomrule
    \end{tabular}
    }
    \caption{FLAN-T5 seq2seq hyper-parameter values. The upper section contains the QLoRA hyper-parameters, and the lower section contains the overall fine-tuning. }
    \label{tab:flan_param}
\end{table}

\begin{table}[h!]
    \centering
    \resizebox{0.75\columnwidth}{!}{
    \begin{tabular}{lc} \toprule
         Hyper-parameter & Value \\ \midrule
         r & 64\\
         $\alpha$ & 128\\
         Dropout & 0.05\\
         Target modules & q\_proj, v\_proj \\
         \midrule
         Max sequence length & 512\\
         Batch size & 2\\
         Gradient accumulation & 4 \\
         Learning rate & $2e-4$\\
         Warm-up steps & 0.03\\
         Scheduler type & cosine\\
         \bottomrule
    \end{tabular}
    }
    \caption{Llama-3-8B hyper-parameter values. The upper section contains the QLoRA hyper-parameters, and the lower section contains the overall fine-tuning. }
    \label{tab:llama_param}
\end{table}

\begin{table}[h!]
    \centering
    \resizebox{0.8\columnwidth}{!}{
    \begin{tabular}{lp{2.5cm}} \toprule
         Hyper-parameter & Value \\ \midrule
         r & 64\\
         $\alpha$ & 16\\
         Dropout & 0.1\\
         Target modules & q\_proj, k\_proj, v\_proj, o\_proj \\
         \midrule
         Max sequence length & 512\\
         Batch size & 2\\
         Gradient accumulation & 2 \\
         Learning rate & $2e-5$\\
         Warm-up steps & 0.03 \\
         Scheduler type & cosine \\
         \bottomrule
    \end{tabular}
    }
    \caption{Tower-7B hyper-parameter values. The upper section contains the QLoRA hyper-parameters, and the lower section contains the overall fine-tuning. }
    \label{tab:tower_param}
\end{table}

\section{Instruction Templates}
\label{inst_templates}
Instruction templates for FLAN-T5, Llama-3 and Tower. The source\_term is the source entry from IATE, the target\_term is the target entry from IATE, source\_language is the source language (i.e. English), target\_id is the target language (i.e. Spanish, German, and Romanian), and glossary\_type is Glossary with one candidate term pair or Glossaries with several candidate terms.

\textbf{FLAN-T5} instruction template for a segment with an identified pair of candidate terms. The prompt is the input for the encoder and the target segment is the input for the decoder:
%\newpage
\begin{lstlisting}
{glossary_type}:
"{source_term}" -> "{target_term}"
...
Translate the source text from {source_id} to {target_id} following the provided translation glossaries.
{source_id}: {source_segment}
\end{lstlisting}

\textbf{FLAN-T5} instruction template with a segment without candidate terms. The prompt is the input for the encoder, and the target segment is the input for the decoder:

\begin{lstlisting}
Translate the source text from {source_id} to {target_id}.
{source_id}: {source_segment}
\end{lstlisting}

\textbf{Llama-3-8B} instruction template for a segment with candidate term pairs:
\begin{lstlisting}
<|begin_of_text|><|start_header_id|>system<|end_header_id|>
You are a helpful translation assistant.<|eot_id|><|start_header_id|>user<|end_header_id|>
{glossary_type}:
"{source_term}" -> "{target_term}"
...
Translate the source text from {source_id} to {target_id} following the provided translation glossaries.
{source_id}: {source_segment}
{target_id}:<|eot_id|>
<|start_header_id|>assistant<|end_header_id|>
{target_segment}<|eot_id|>  
\end{lstlisting}

\textbf{Llama-3-8B} instruction template for a segment without candidate term pairs:
\begin{lstlisting}
<|begin_of_text|><|start_header_id|>system<|end_header_id|>
You are a helpful translation assistant.<|eot_id|><|start_header_id|>user<|end_header_id|>
Translate the source text from {source_id} to {target_id}.
{source_id}: {source_segment}
{target_id}:<|eot_id|>
<|start_header_id|>assistant<|end_header_id|>
{target_segment}<|eot_id|>  
\end{lstlisting}

\textbf{Tower-7B} instruction template for a segment with candidate term pairs:
\begin{lstlisting}
<|im_start|>user
{glossary_type}:
"{source_term}" -> "{target_term}"
...
Translate the source text from {source_id} to {target_id} following the provided translation glossaries.
{source_id}: {source_segment}
{target_id}:<|im_end|>
<|im_start|>assistant
{target_segment}<|im end|>
\end{lstlisting}

\textbf{Tower-7B} instruction template for a segment without candidate term pairs:
\begin{lstlisting}
<|im_start|>user
Translate the source text from {source_id} to {target_id}.
{source_id}: {source_segment}
{target_id}:<|im_end|>
<|im_start|>assistant
{target_segment}<|im end|>
\end{lstlisting}

\end{document}